\documentclass[conference]{IEEEtran}
\IEEEoverridecommandlockouts

\usepackage{cite}
\usepackage{amsmath,amssymb,amsfonts}
\usepackage{algorithmic}
\usepackage{textcomp}
\usepackage{float}
\usepackage{booktabs}
\usepackage[table,xcdraw]{xcolor}
\usepackage{subcaption}
\usepackage{url}
\usepackage[export]{adjustbox} 
\usepackage{xcolor}
\usepackage{dblfloatfix}
\def\BibTeX{{\rm B\kern-.05em{\sc i\kern-.025em b}\kern-.08em
    T\kern-.1667em\lower.7ex\hbox{E}\kern-.125emX}}
\usepackage{flushend}

\begin{document}

\title{EvSegSNN: Neuromorphic Semantic Segmentation for Event Data

}

\author{\IEEEauthorblockN{Anonymous Authors}}

\author{\IEEEauthorblockN{1\textsuperscript{st} author}
\IEEEauthorblockA{\textit{Hidden for blind peer review} \\
\textit{Hidden for blind peer review}\\
Hidden for blind peer review \\
Hidden for blind peer review}
\and
\IEEEauthorblockN{2\textsuperscript{nd} author}
\IEEEauthorblockA{\textit{Hidden for blind peer review} \\
\textit{Hidden for blind peer review}\\
Hidden for blind peer review \\
Hidden for blind peer review}}

\author{\IEEEauthorblockN{Dalia Hareb}
\IEEEauthorblockA{\textit{i3S / CNRS} \\
\textit{Université Côte d'Azur}\\
Sophia Antipolis, France \\
hareb@i3s.unice.fr}
\and
\IEEEauthorblockN{Jean Martinet}
\IEEEauthorblockA{\textit{i3S / CNRS} \\
\textit{Université Côte d'Azur}\\
Sophia Antipolis, France \\
jean.martinet@univ-cotedazur.fr }
}

\maketitle

\begin{abstract}
Semantic segmentation is an important computer vision task, particularly for scene understanding and navigation of autonomous vehicles and UAVs. Several variations of deep neural network architectures have been designed to tackle this task. However, due to their huge computational costs and their high memory consumption, these models are not meant to be deployed on resource-constrained systems. To address this limitation, we introduce an end-to-end biologically inspired semantic segmentation approach by combining Spiking Neural Networks (SNNs, a low-power alternative to classical neural networks) with event cameras whose output data can directly feed these neural network inputs. We have designed EvSegSNN, a biologically plausible encoder-decoder U-shaped architecture relying on Parametric Leaky Integrate and Fire neurons in an objective to trade-off resource usage against performance. The experiments conducted on DDD17 demonstrate that EvSegSNN outperforms the closest state-of-the-art model in terms of MIoU while reducing the number of parameters by a factor of $1.6$ and sparing a batch normalization stage.
\end{abstract}


\section{Introduction}
\begin{figure}[t]
    \centering
    \includegraphics[scale=0.26] {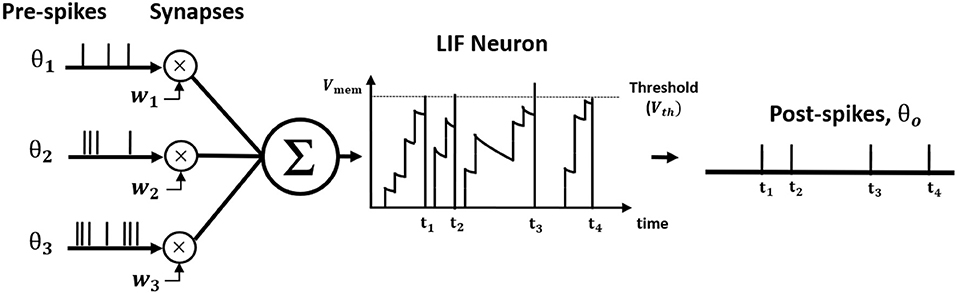}
    \caption{LIF neuron (Image taken from \cite{lee2020enabling}).}
    \label{fig:lif_neuron}
\end{figure}

Neuromorphic computing guided by the principles of biological neural computing and inspired by the human brain's interaction with the world has been studied for years to open up the possibility of extending the use of artificial neural networks from hardware with huge computational costs and high memory consumption to embedded systems equipped with power-constrained components such as Internet of Things (IoT) devices, automotive and AR/VR. In this sense, models like Spiking Neural Networks (SNNs) are strongly represented, because they take an additional level of inspiration from biological neural systems compared to standard Artificial Neural Networks (ANN) by integrating the concept of time into their operating model \cite{cordone2022object}.
\\
These neural models are based on timestamped discrete events called spikes. A commonly used neuron model (activation function) for SNNs is the Leaky-Integrate-and-fire (LIF) neuron (Fig. \ref{fig:lif_neuron}). In a discrete formulation, the membrane potential $V_i^t$ is measured by summing up the decayed membrane potential from previous timesteps $V_i^{t-1}$ with a sum of the weighted spike signals $W_{ij}$ triggered by the pre-synaptic neurons $j$:
\begin{equation}
    V_i^t = \lambda V_i^{t-1} + \sum_j W_{ij} \cdot \theta_j^t
\end{equation}

\noindent When this membrane potential $V_i^t$ exceeds the firing threshold $V_{th}$, a spike $S_i^t$ is generated by the neuron $i$:
\begin{equation} \label{spike}
    \theta_i^t = \left\{
    \begin{array}{ll}
        1 & \mbox{if  }  V_i^t > V_{th} \\
        0 &\mbox{otherwise} 
    \end{array}
\right.
\end{equation}
After a spike is emitted, the $V_i^t$ value is either lowered by the amount of the threshold (soft reset) or reset to the minimum voltage such as zero (hard reset).

Besides taking advantage of low power consumption as they consume energy only when spikes are triggered and low computation latency, due to the asynchronous computation of the spikes and the speed of their spread \cite{pfeiffer2018deep}. SNNs are the most suitable interface for event-based sensors such as event cameras.

Event cameras are bio-inspired sensors designed to overcome the standard frame cameras' drawbacks. Instead of recording the video as a sequence of dense frames with every image collected with a constant rate \cite{viale2021carsnn}, they asynchronously measure per-pixel brightness changes and output a stream of $ <x,y,p,t> $ events that encode the time $t$, location $(x, y)$ and sign of the brightness changes, named polarity $p$ (positive if the pixel brightness increases, negative otherwise).
\\
Thus, thanks to their structure, these offer multiple advantages over standard cameras\cite{gallego2020event}, such as a high temporal resolution (in the order of $\mu s$), very high dynamic range ( $>120$ $dB$ vs. $60$ $dB$ for frame-based cameras), a low latency ($10$ $ms$ on the lab bench, and sub-millisecond in the real world) and low power consumption ($100$ $mW$ max). In addition, they are extremely useful when coupled with the SNNs, because the generated events/spikes can directly feed the SNNs’ inputs.

In this paper, we introduce EvSegSNN, a spiking Convolutional Neural Network (CNN) combined with event data to tackle the semantic segmentation problem.
%
This computer vision task is mostly used in self-driving cars and UAVs that require real-time processing with reduced energy consumption.
The proposed approach is biologically plausible, namely because it relies on SNNs. More importantly, it uses max-pooling, concatenation and does not require batch normalization \cite{ioffe2015batch}, which is a standard process used in deep learning to fix the means and variances of network layers by applying on a mini-batch $(x_1, x_2,...,x_i,...,x_m)$ of size $m$ the following equation:
\begin{equation}
    BN_{\gamma, \beta}(x_i) = \gamma x'_i + \beta
\end{equation}
where $\gamma$ and $\beta$ indicate the parameters to be learned and $x'_i$ is:
\begin{equation}
     x'_i = \frac{x_i - \mu}{\sqrt{\sigma^2 + \epsilon}}
\end{equation}
Here $\epsilon$ is a constant added to the mini-batch variance for numerical stability, $\mu$ and $\sigma^2$ represent the mini-batch mean and the mini-batch variance, respectively. They are measured as follows:
\begin{equation}
    \mu = \frac{1}{m} \sum_{i=1}^m x_i
\end{equation}

\begin{equation}
     \sigma^2 =\frac{1}{m} \sum_{i=1}^m(x_i - \mu)^2
\end{equation}

The proposed model is validated with the DDD17 dataset, an event dataset featuring driving sequences. Our main contributions are:
\begin{enumerate}
    \item We design a spiking light Unet model for semantic segmentation that outperforms the state-of-the-art model with $5.58\%$ of absolute MIoU while reducing the number of parameters by $62\%$.
    \item We successfully train a large spiking neural network using Surrogate Gradient Learning.     
    \item We propose a biologically plausible spiking neural network Unet model that does not require batch normalization in any layer.
    
\end{enumerate}

The paper is organized as follows. In section \ref{related_work}, we introduce some research works that inspired our contribution. In section \ref{contribution}, we explain our methodology by describing our network architecture including the main differences with the state-of-the-art model we take as a baseline \cite{kim2022beyond}. Then, in section \ref{experiments}, we present the data pre-processing, the implementation details, and the performance results in terms of MIoU, accuracy, and number of parameters to compare our EvSegSNN model to the state-of-the-art model with and without batch normalization on the one hand, and to the original Unet designed for semantic segmentation on the other.

\section{Related work} \label{related_work}
\begin{figure}[!b]
    \centering
    \includegraphics[scale = 0.3]{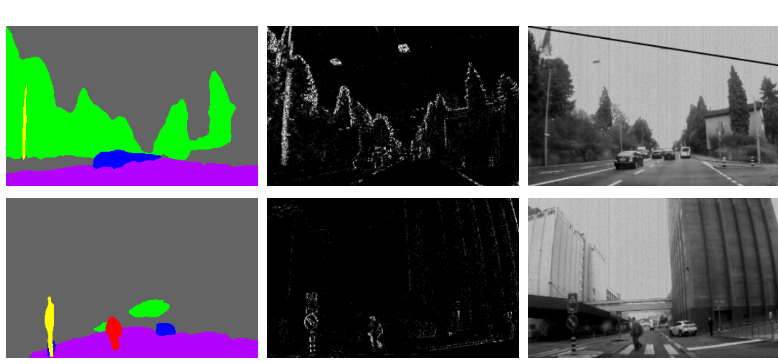}
    \caption{Semantic segmentation results (left) of accumulated events within an interval of $50ms$ (middle). Grayscale images represent the scenes captured by the event camera (right). Image from~\cite{alonso2019ev}.}
    \label{seg_evsegnet}
\end{figure}

Semantic segmentation is a computer vision task that assigns a label, corresponding to a given class, for each pixel in the image (Fig. \ref{seg_evsegnet}). In recent years, this challenging task has been resolved using deep learning approaches based on different variations of encoder-decoder CNN architectures. The encoder downsamples the image given as input and the decoder upsamples the result returned by the encoder until it reaches the original size of the image. Among these techniques, we have Fully Convolutional Network (FCN) \cite{long2015fully}, a model obtained after transforming the fully connected layers in the most adapted classification networks (AlexNet \cite{krizhevsky2017imagenet}, VGG16 \cite{simonyan2014very} and GoogLeNet \cite{szegedy2015going}) into fully convolutional ones to output spatial maps instead of classification scores. SegNet \cite{badrinarayanan2017segnet} consisting of a pre-trained VGG16 without the fully connected layers as encoder and a decoder that aims to upsample the image to its original size using the max pooling indices sent by the encoder. Unet \cite{ronneberger2015u}, a U-shaped encoder-decoder network architecture, which consists of four encoder blocks that half the spatial dimensions and double the number of filters and four decoder blocks that do the opposite process, meaning it doubles the spatial dimensions and half the number of feature channels. Both are connected via skip connections. And Xception \cite{chen2018encoder} which stands for “Extreme Inception”, a $36$ convolutional layers split in $3$ flows such that the input, first, goes through the 1st flow, then through the 2nd flow which is repeated eight times, and goes finally through the last exit flow. 
As input data, all these previous works have used only images captured by classic cameras. The first research work exploring the use of event-based cameras to do semantic segmentation we can find is Alonso et al. \cite{alonso2019ev}. Their model consists of an encoder-decoder architecture that has been inspired by current state-of-the-art semantic segmentation CNNs, slightly adapted to use the event data encoding. An encoder represented by Xception model in which all the training is concentrated, and a light decoder connected to the encoder via skip connections to help deep neural architecture avoid the vanishing gradient problem and to help the decoder build an image using the fine-grained details learned in the encoder.

However, since event-driven cameras produce spikes asynchronously, their data are more suitable as input for a Spiking Neural Network which also works with spikes asynchronously. So far, this combination has been used for fundamental computer vision tasks like image classification \cite{ceolini2020hand}, \cite{taunyazov2020event}. To the best of our knowledge, the only works that extend its use to semantic segmentation are \cite{zhang2023energy} and \cite{kim2022beyond}. Zhang et al. \cite{zhang2023energy} introduced a Spiking Context Guided Network composed of SCG blocks, including membrane shortcut blocks encompassing $1 \times 1$ convolutions for transformation, $3 \times 3$ standard and dilated convolutions whose respective outputs are fused using batch normalization and concatenation. Although the SCG module is fully spiking, the subsequent global context extractor, based on channel-wise attention to extract the global context of the image, relies on non-spiking computation. This is evident in the use of global average pooling returning floats, followed by two fully connected layers with ReLU and sigmoid activation functions, as opposed to spiking neurons. Consequently, the network is not a Spiking but a hybrid network.
In a different approach, Kim et al. \cite{kim2022beyond} designed a spiking Fully Convolutional Network that consists of an encoder-decoder architecture connected via skip connections using for each convolution layer Batch-Normalization-Through-Time (BNTT) \cite{kim2020revisiting} that is a batch normalization applied at every timestep. In their experiments, they use LIF neurons and piece-wise function during the forward and backward propagation, respectively with a soft reset scheme. As input, they use re-scaled image resolution of $346 \times 200$ pixels to $64 \times 64$ pixel of data provided by \cite{alonso2019ev}, without the mean and standard deviation channels measured by computing the arithmetic mean and standard deviation of the normalized timestamps of events happening at each pixel $(x_i, y_i)$ within an interval of $50ms$, computed separately for the positive and negative events. As a result, their model reaches good performances in terms of MIoU, however, the use of batch normalization, the addition of the feature maps and average pooling make their spiking model non-bio-plausible \cite{shaw2020biological} as the model is not fully spiky. Nevertheless, this work serves as a reference due to its proximity to our idea.

\section{EvSegSNN for semantic segmentation of event data} \label{contribution}
\begin{figure*}[t!]
  \vspace{-0.3cm}
  \centering
  \begin{subfigure}[t]{0.03\textwidth}\textbf{A}\end{subfigure}   
    \begin{subfigure}{.41\textwidth}
        \includegraphics[width=1\linewidth, height=1.1\linewidth, valign=t]{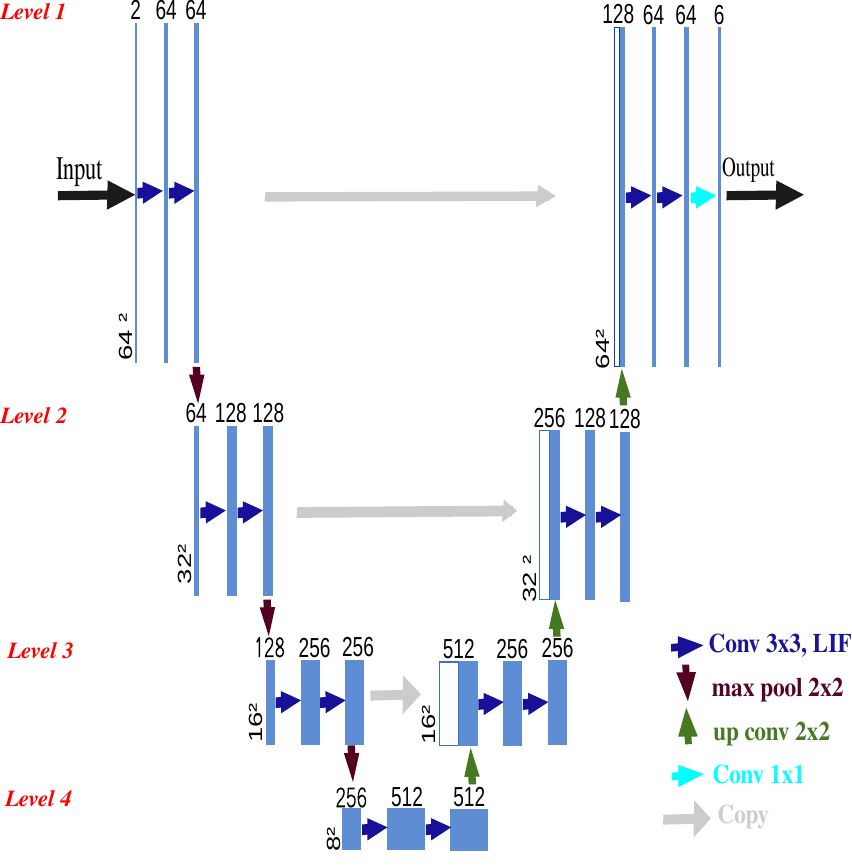}
    \end{subfigure}
    \hfill
    \begin{subfigure}[t]{0.03\textwidth}\textbf{B}\end{subfigure}
    \begin{subfigure}{.51\textwidth}
        \includegraphics[width = 1\linewidth, height=1\linewidth,  valign = t]{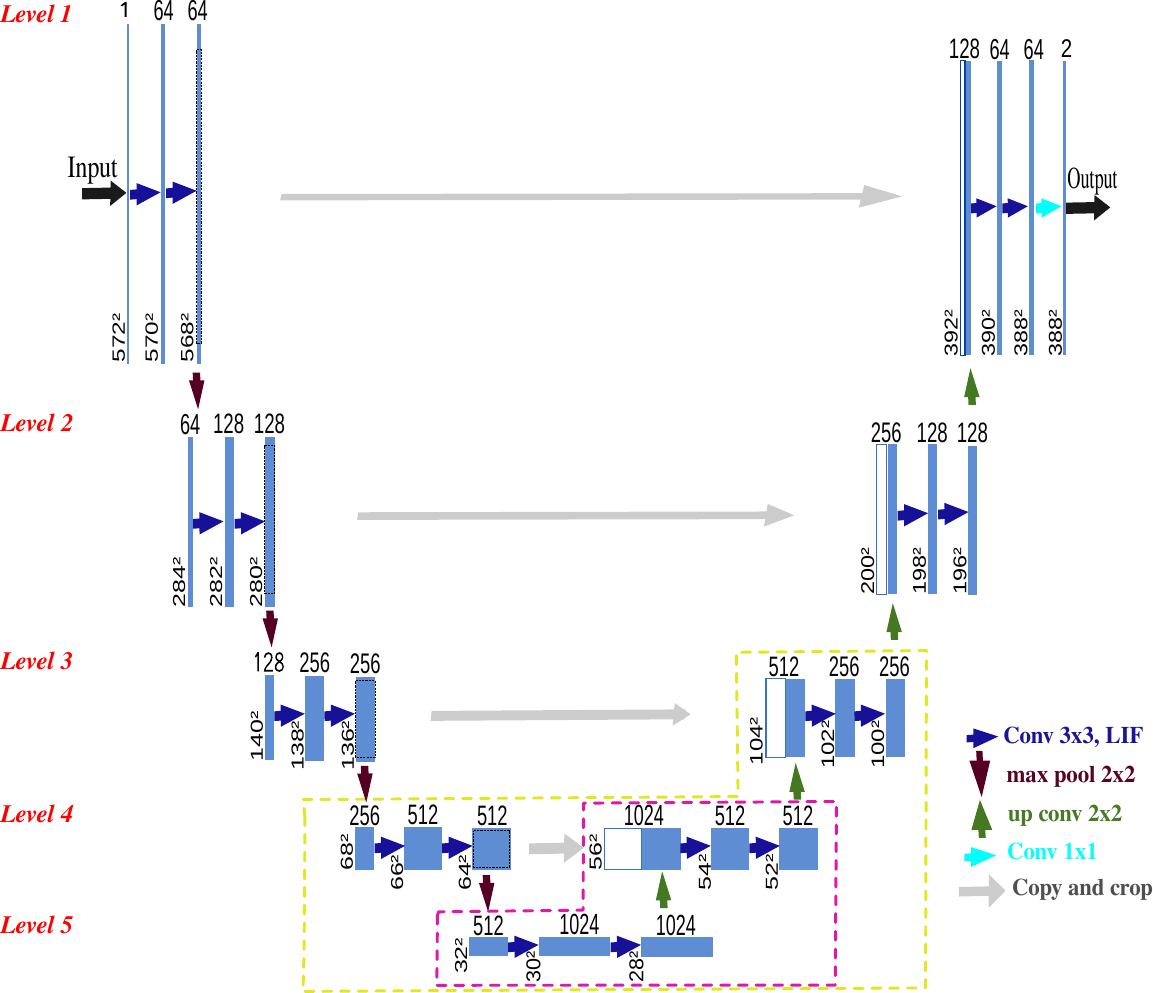}
    \end{subfigure}
    
    \caption{\textbf{A}: The proposed EvSegSNN was obtained after reducing the Unet size and its depth by one level corresponding to the layers within the pink box. \textbf{B}: The original Unet model \cite{ronneberger2015u}. The pink and yellow boxes correspond to 2 depth levels reduced to obtain 2 light Unet models.}
  \label{fig:results}
  \vspace{-0.2cm}
\end{figure*}

Our work is grounded on a Unet model \cite{ronneberger2015u} adapted for Spiking Neural Network (Fig. \ref{fig:results}B). To reduce its complexity, inspired by \cite{sunn2022light} who proposed a light Unet model consisting of a smaller Unet with fewer parameters than the original, we design a light Unet topology EvSegSNN with four depth levels, as illustrated in Fig. \ref{fig:results}A.

Compared to the original Unet, our light Unet model spares 4 convolution layers, 1 max-pooling layer, and 1 upsampling layer -- represented within the pink dashed box of Fig. \ref{fig:results}B. This design choice has been made for two main reasons: (1) the deepest layer represents the heaviest part of Unet in terms of the number of parameters, and (2) the deepest Unet layer in the pink dashed box is particularly suited for dense 3-channel RGB data, while event data is sparse and it features only two polarities. Hence, by lowering the number of layers in the encoder whose role is to compress the input through the layers to keep the most important details, no performance drop can be noticed in the event frames processing. In addition, its size has been reduced to obtain a dimension equal to $64\times64$ in the first layer.

To preserve the biological plausibility of Spiking Neural Networks (SNNs), our proposed model diverges from normalization practices. In contrast to our baseline model \cite{kim2022beyond}, which employs the Batch-Normalization-Through-Time technique at each convolutional layer, our model refrains from incorporating any normalization form. Additionally, our model utilizes feature map addition instead of concatenation, avoiding the transformation of binary values to integers exceeding 1. Furthermore, our approach incorporates max-pooling as opposed to average pooling, ensuring the preservation of binary values rather than returning floating-point numbers. This strategic departure from normalization and related techniques is instrumental in maintaining the biological plausibility of our SNN model.

\begin{figure}[t!]
    \centering
    \includegraphics[scale = 0.21]{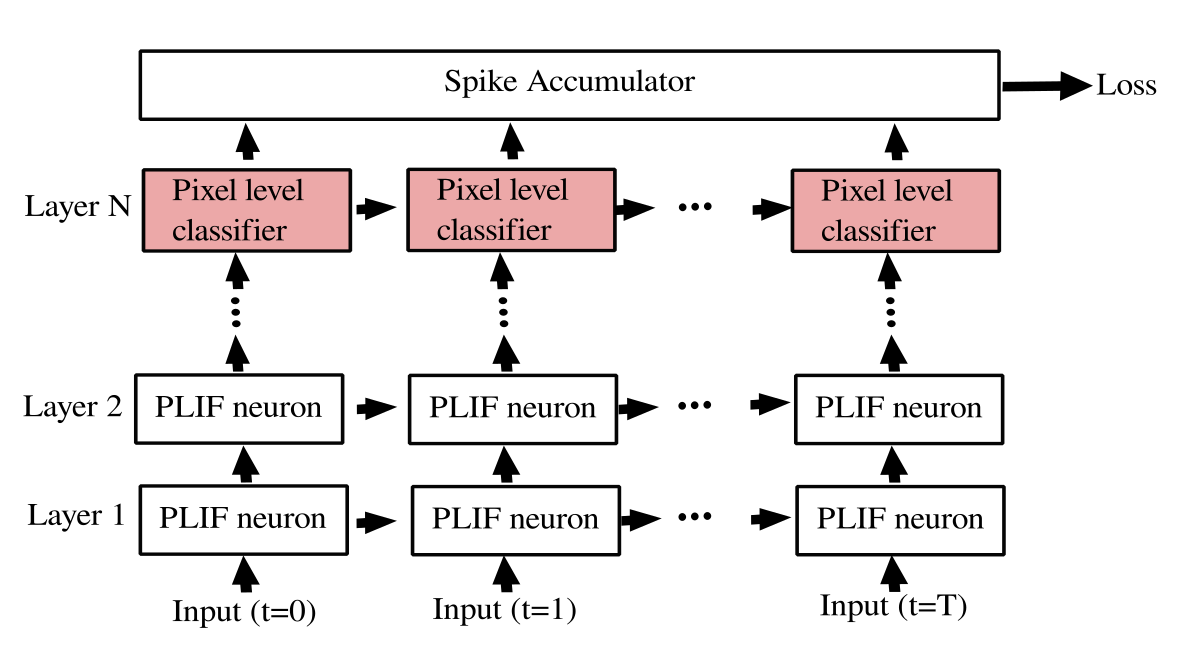}
    \caption{Computational graph of both encoder decoder parts unrolled over multiple timesteps. Adapted from \cite{kim2022beyond}.}
    \label{fig:youngeun}
\end{figure}

The computing graph of the model is described in Fig. \ref{fig:youngeun} showing that each neuron's membrane potential is measured by combining the previous membrane potential (horizontal lines) and the previous layer (vertical lines). The output spikes across all timesteps are accumulated to generate a 2-dimensional probabilistic map in the last layer whose number of channels is equal to the number of segmentation classes and whose activation function is set to \textit{None}.

More particularly, our model is made up of Parametric LIF (PLIF) neurons introduced by \cite{fang2021incorporating}, which are known to be less sensitive to the initial values of the learnable time constants compared to LIF. Also, it uses a soft-reset scheme because, unlike a hard reset, this scheme helps retain the residual information after the reset, thus avoiding information loss. During its training, because of the behaviour of PLIF neurons making the gradient of their output spikes with respect to their membrane potential non-differentiable, the training optimization via back-propagation of the loss is performed based on surrogate Back-propagation Through Time (BPTT): 
\begin{itemize}
    \item \textit{Surrogate Back-propagation} refers to the use of the backward gradient approximation using the piece-wise quadratic surrogate spiking function defined in Eq. \ref{surrogate} and taken from \cite{SpikingJelly} documentation.
    \begin{equation} \label{surrogate}
        \frac{\delta S}{\delta V}  = \left\{
        \begin{array}{ll}
            0 & |V|> \frac{1}{\alpha} \\
            - \alpha^2|V|+\alpha & V \leq \frac{1}{\alpha }
        \end{array}
    \right.
    \end{equation}
    \item \textit{Through Time} refers to the accumulated gradients over all timesteps across all layers.      
\end{itemize} 

Where $S$, $V$, and $\alpha$ represent the output spike, the membrane potential, and a scalar parameter to control the smoothness of the gradient, respectively.
\\
The loss $L$ is a pixel-wise classifier loss that is measured by summing the error between the estimated pixels' classes and the ground-truth:

\begin{equation} \label{crossentropy}
    L = -\frac{1}{N} \sum_{i=1}^{H} \sum_{j=1}^{W}  \sum_{k=1}^{C} y_{i,j} log \left( \frac{e^{V_{i,j}^k}}{\sum_{m=1}^{C} e^{V_{i,j}^m} }\right)
\end{equation}
Here, $N$ is the size of the minibatch, $C$ is the number of classes, and $H$, $W$ represent the height and the width of the pseudo-frame, respectively. $y_{i,j}$ is the ground-truth label, and $V_{i,j}^m$ stands for the membrane potential of the neurons corresponding to the pixel $(i,j)$ in the last layer.

\section{Experimental validation} \label{experiments}
\subsection{Event-based dataset} \label{evsegnet_dataset} 
\begin{figure*}[t!] 
  \centering  
    \begin{subfigure}{.32\textwidth}
        \includegraphics[width = 1.00\linewidth, height=2.3\linewidth]{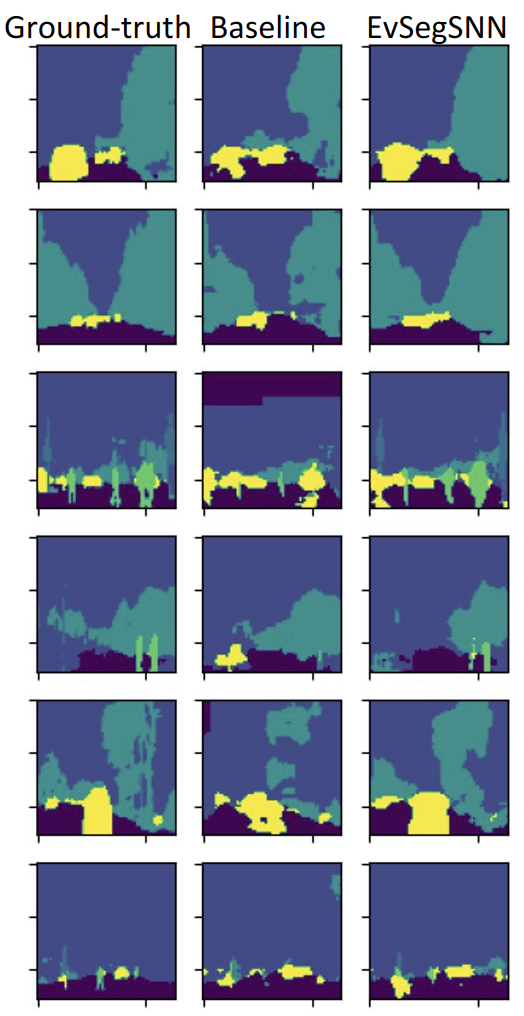}
    \end{subfigure}
    \hfill
    \begin{subfigure}{.32\textwidth}
        \includegraphics[width=1\linewidth, height=2.3\linewidth]{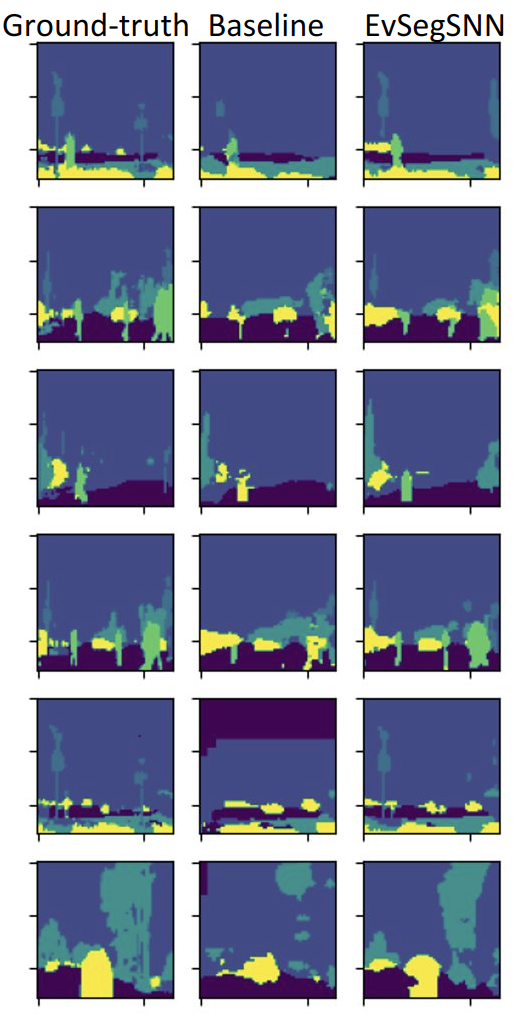}
    \end{subfigure}
    \hfill
    \begin{subfigure}{.32\textwidth}
        \includegraphics[width=1\linewidth, height=2.3\linewidth]{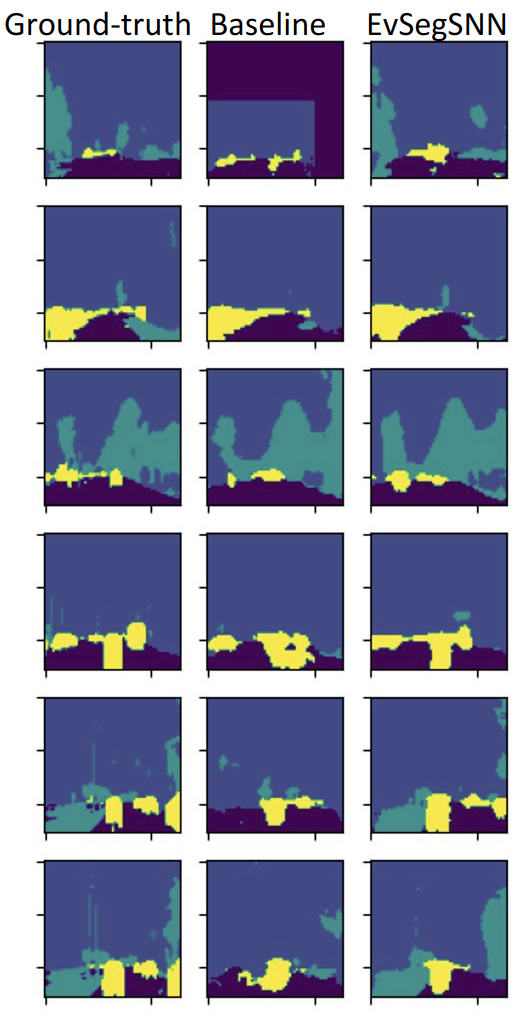}
    \end{subfigure}
    \caption{ Example of semantic segmentation results using Light Unet on the DDD17. We visualize the Ground-Truth - Baseline prediction - EvSegSNN prediction }
  \label{fig:results_seg}  
\end{figure*}

The DAVIS Driving Dataset 2017 (DDD17) \cite{binas2017ddd17} contains 40 different driving sequences of event data captured by an event camera. Since the original dataset provides only grey-scale images and event data without semantic segmentation labels, we used the segmentation labels provided by Alonso et al. \cite{alonso2019ev} who generated an approximated set of labels using a CNN model trained with the Cityscapes dataset. These labels are not as perfect as if it's manually annotated and this can be noticed in the left column of each of the 3 samples illustrated in Fig. \ref{fig:results_seg}.
This dataset includes 20 different sequence intervals taken from 6 of the original DDD17 sequences. Furthermore, since only multi-channel representations of the events (normalized sum, mean, and standard deviation for each polarity) are made available, we extracted the original events from DDD17 with the original $<x,y,p,t>$ structure using DDD20 tools\footnote{DDD tools: https://github.com/SensorsINI/ddd20-utils.} and selected events corresponding to the frames with a ground-truth. 

We have used a $5$-fold cross-validation strategy in our experiments with $80\%$-$20\%$ train-test split. This choice has been made since the original split from \cite{kim2022beyond} and \cite{alonso2019ev} revealed a significant difference between train and test performances corresponding to a MIoU equal to $63.91\%$ and $32.92\%$, respectively.

\subsection{Implementation}
\subsubsection{Hyperparameters}
For the experimental validation, we first implemented Kim's et al. \cite{kim2022beyond} model based on the open source code\footnote{https://github.com/Intelligent-Computing-Lab-Yale/BNTT-Batch-Normalization-Through-Time.} associated to their research work \cite{kim2020revisiting} related to Batch-Normalization-Throught-Time. The hyperparameters of the model are available in their paper\cite{kim2022beyond}.

We implemented our model using SpikingJelly \cite{SpikingJelly}, a Python framework based on Pytorch dedicated to SNNs. We trained the model on a desktop computer equipped with an NVIDIA RTX A5000 GPU card using the hyperparameters given in Table \ref{tab:hyperprmt}.
\begin{table}[t!]
\centering
\caption{EvSegSNN hyperparameters}
\label{tab:hyperprmt}
\begin{tabular}{@{}|l|c|@{}}
\hline
Optimizer                                                     & Nadam           \\ 
Learning rate                                                 & $2e-3$          \\  
Learning rate scheduling                                      & $[8, 16, 24, 50]$ with a factor of 10 \\ \hline
Batch size                                                    & $16$            \\ 
Number of epochs                                              & $70$            \\ 
Timesteps                                                     & $20$            \\ \hline
Leak factor                                                   & $0.99$          \\ 
Membrane threshold    & $1.0$          \\ \hline
\end{tabular}
\end{table}

\subsubsection{Performance metrics}
To evaluate our model and compare it to the baseline, we use standard metrics of semantic segmentation as used by Alonso et al. \cite{alonso2019ev} i.e. \textit{accuracy} and \textit{Mean Intersection over Union (MIoU)}:
\begin{equation} \label{accuracy}
 \begin{aligned}
    accuracy(y, \widehat{y}) &= \frac{1}{N} \sum_{i=1}^{N} \delta(y_i, \widehat y_i) \\
    &= \frac{TP + TN }{TP+TN+FP+FN}
\end{aligned}   
\end{equation}
\begin{equation} \label{miou}
 \begin{aligned}
      MIoU(y, \widehat y) & = \frac{1}{C} \sum_{j=1}^{C} \frac{\sum_{i=1}^{N} \delta(y_{i,c},1)\delta(y_{i,c}, \widehat{y_{i,c}} )}{\sum_{i=1}^{N} max(1, \delta(y_{i,c},1)\delta(\widehat{y_{i,c}}, 1) )}  \\
      &= \frac{TP}{TP+TN+FP+FN}
 \end{aligned}   
\end{equation}
In Eq. \ref{accuracy} and Eq. \ref{miou}, $y$ and $\widehat{y}$ are the desired output and the system output respectively, $C$ is the number of classes, $N$ is the number of pixels and $\delta $ denotes the Kronecker delta function, TP, TN, FP, and FN stand for true positive, true negative, false positive and false negative, respectively.

Moreover, we compare our model to the baseline in terms of \textit{number of parameters} whose reduction involves a reduction of the runtime as well as the energy consumption, which represents the first motivation for using SNNs.

\subsection{Performance results} \label{performance_results}

\begin{figure}[t!]
    \centering
    \includegraphics[scale=0.33] {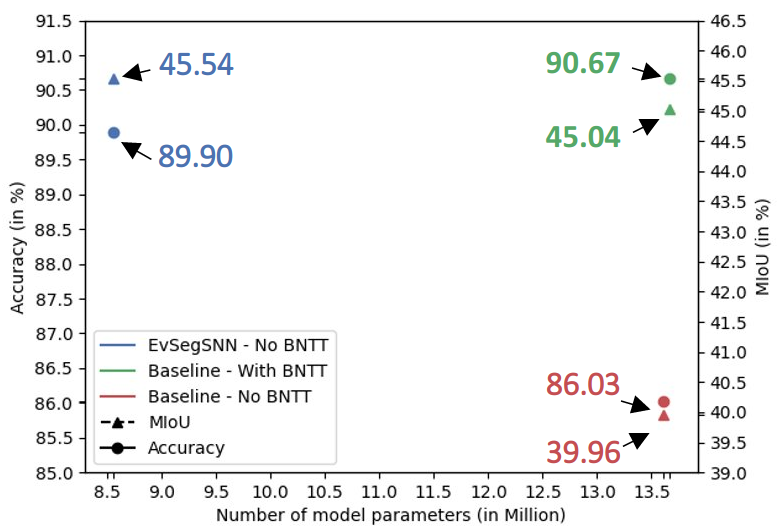}
    \caption{MIoU and accuracy of EvSegSNN and the baseline (Kim et al. \cite{kim2022beyond}) with and without BNTT w.r.t. the number of parameters.}
    \label{fig:mine_kim}
\end{figure}

Figure \ref{fig:mine_kim} shows a performance comparison of the proposed EvSegSNN (blue) and the baseline model with and without Batch-Normalization-Through-Time (green and red, respectively) in terms of MIoU and accuracy, with respect to the number of parameters.

It is interesting to note that, on the one hand, even though the number of parameters of our model ($8.55M$) is reduced by a factor of $1.6$ compared to Kim et al. \cite{kim2022beyond}, the performances of both models remain very close. On the other hand, considering the bio-plausibility of Spiking Neural Networks, we notice that EvSegSNN which is bio-plausible, as it is built with max-pooling and concatenation and without BNTT,  yields a MIoU of $45.54\%$ and accuracy of $89.90\%$, while the baseline designed without BNTT (while keeping the average pooling and the addition of feature maps) only obtains a MIoU of $39.96\%$ and accuracy of $86.03\%$. Therefore, in addition to being bio-plausible, EvSegSNN outperforms the baseline model without BNTT while having a number of parameters $1.6$ times smaller.
Note that the number of parameters of Kim et al. \cite{kim2022beyond} model without BNTT is slightly reduced compared to the original one (with BNTT) due to the removal of trainable parameters used for BNTT.
By taking a close look at MIoU results of EvSegSNN during training and testing summarized in Table \ref{tab:my-table}, we can see that the MIoU of the \textit{Background}, the \textit{Road}, the \textit{Tree} and the \textit{Car} classes show good performances, while the \textit{Panel} and the \textit{Pedestrian} classes show unsatisfying values. This can be explained by the average number of pixels belonging to each class in all the DDD17 frames. Indeed, in Table \ref{tab:number_pixels}, we can see that the four first classes are represented by a large number of pixels while the \textit{Panel} and \textit{Pedestrian} classes contain on average less than 500 and 100 pixels respectively, thus explaining the poor performances for both these classes.

\begin{table}[b!]
\centering
\caption{Class-wise MIoU results for EvSegSNN.}
\label{tab:my-table}
\begin{tabular}{@{}|l|c|c|c|c|c|c|@{}}
\hline
           &   \scriptsize{Road}  &   \scriptsize{Background} &   \scriptsize{Panels} &   \scriptsize{Tree}  &   \scriptsize{Pedestrian} &   \scriptsize{Car}   \\ \hline
  \scriptsize{MIoU Train} & \scriptsize{83.93} & \scriptsize{93.01}      & \scriptsize{\textbf{17.26}}  & \scriptsize{75.92} & \scriptsize{\textbf{14.81} }     & \scriptsize{72.75} \\ \hline
  \scriptsize{MIoU Test}  & \scriptsize{79.14} & \scriptsize{89.26}      & \scriptsize{\textbf{6.25}}   & \scriptsize{48.54} &\scriptsize{\textbf{2.44}}       & \scriptsize{47.63} \\ \hline
\end{tabular}
\end{table}

\begin{table}[b!]
\centering
\caption{Average number of pixels per class in DDD17 }
\label{tab:number_pixels}
\begin{tabular}{|l|c|c|c|c|c|c|}
\hline
\multicolumn{1}{|c|}{} &  \scriptsize{Road}  &   \scriptsize{Background} &   \scriptsize{Panels} &   \scriptsize{Tree}  &   \scriptsize{Pedestrian} &   \scriptsize{Car}  \\ \hline
  \scriptsize{Pixels} &\scriptsize{7.49K} & \scriptsize{47.22K} &  \scriptsize{\textbf{0.476K}} &  \scriptsize{10.55K} &  \scriptsize{\textbf{0.087K}} &  \scriptsize{3.36K}\\ \hline
\end{tabular}
\end{table}

\begin{figure}[t!]
    \centering
    \includegraphics[width=25em, height=20em] {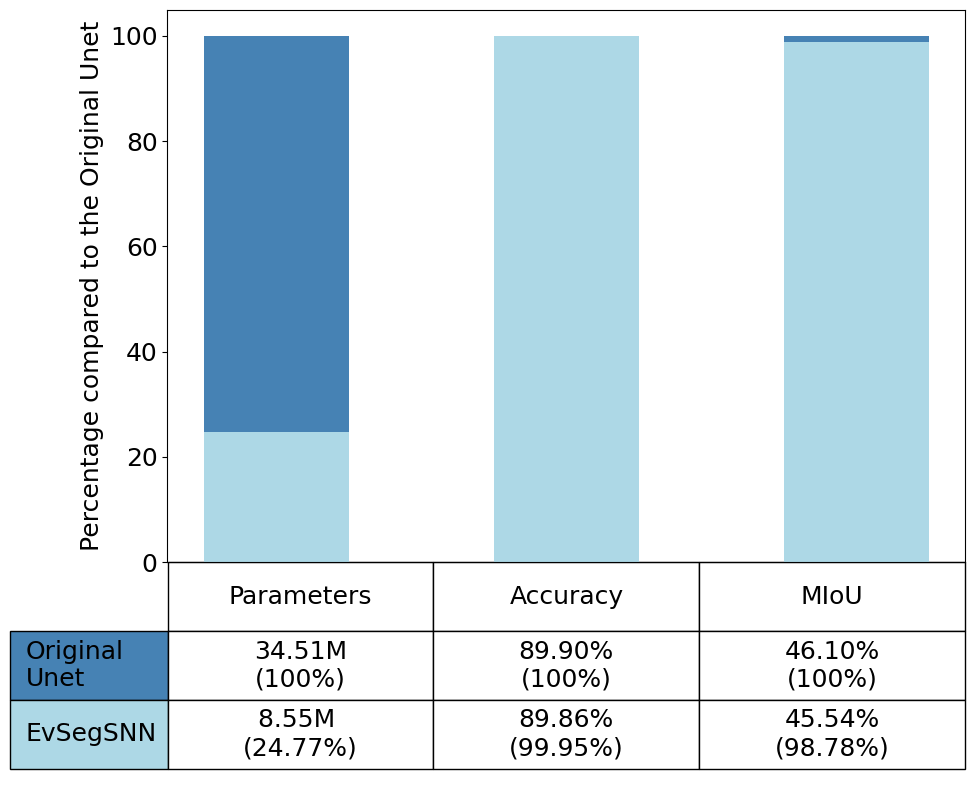}
    \caption{Comparison of EvSegSNN to the Original Unet.}
    \label{fig:unets}
\end{figure}

Figure \ref{fig:unets} presents a comparison of the proposed model with the original Unet in terms of the number of parameters and performances. The original Unet is taken as a reference model for EvSegSNN with the maximal set of parameters, thus scoring $100\%$ in each metric. Note that, similarly to EvSegSNN, the input size of the original Unet has also been reduced to fit the $64\times64$ size of the dataset.

As we can see, by designing EvSegSNN whose number of trainable parameters is equal to $8.55M$ (about $25\%$ the number of parameters of the original Unet), we obtain performances in terms of MIoU and accuracy almost equal to the original. This result shows that we successfully designed a smaller and less complex Unet while maintaining the performance. Thus, this also shows that our initial hypothesis stating that due to its structure, event data does not require a deep encoder to preserve the most important details while being down-sampled is verified.

Note that if we go further in the simplification and introduce another light Unet with much fewer parameters ($2M$) by sparing the layers represented within the yellow dashed box of the Fig. \ref{fig:results}B, we notice a significant decrease of the semantic segmentation quality whose evaluation metrics return an MIoU of $40.19\%$ and an accuracy of $84.67\%$ representing a drop of $5.91\%$ and $5.23\%$ respectively, compared to the original Unet.

\subsection{Limitations} \label{performance_results}
Although the proposed model offers several advantages, it is important to acknowledge its limitations. Specifically, the weighted average firing rate of EvSegSNN, as calculated according to Eq. \ref{firing_rate}, is observed to be 100x higher than that of Kim et al.'s model \cite{kim2022beyond}. This discrepancy can be attributed to two primary factors:
\begin{enumerate}
    \item EvSegSNN has a significantly greater number of neurons, amounting to $3.8M$, which is 3x greater than the $1.3M$ neurons present in the baseline model \cite{kim2022beyond}.
    \item The mean spiking rate per neuron per layer and per time step (as given by Eq. \ref{firing_rate_layer}) is, on average, 10x greater than that of the baseline.  
\end{enumerate}

The firing rate of the model is defined as:
\begin{equation}\label{firing_rate}
    FR_{model} = \frac{1}{L}\sum_{l=1}^L N_l\times FR_{layer}(l)    
\end{equation}
where $L$ and $N_l$ respectively refer to the number of layers and the number of neurons in layer $l$. The average firing rate $FR_{layer}(l)$ of a layer $l$ is defined as follows:
\begin{equation} \label{firing_rate_layer}
    FR_{layer}(l) = \frac{1}{MT}\sum_{i=1}^M\sum_{t=1}^T  \frac{NbS^l_{i,t}}{H_lW_l}
\end{equation}
Here, $M$ represents the total number of samples, and $T$ denotes the number of time steps. $NbS^l_{i,t}$ refers to the number of spikes observed in layer $l$ with dimensions $H_lW_l$, during time step $t$ while segmenting the pseudo-frame $i$.

As stated in Section \ref{experiments}, we employed the PyTorch framework to implement the baseline model from scratch, drawing inspiration from Kim et al.'s earlier work \cite{kim2020revisiting}. Meanwhile, our model was implemented using SpikingJelly, representing an intermediate layer between our code and PyTorch. This difference, coupled with the higher number of neurons used in our model, resulted in a 2x the time inference per sample in our model compared to the baseline.

Although our model has certain limitations, it is still possible to optimize, quantize and then implement it on specialized neuromorphic hardware, such as Intel Loihi \cite{davies2018loihi} and IBM TrueNorth \cite{akopyan2015truenorth}, to take advantage of binary sparse spikes that flow through the network. In contrast, a neuromorphic hardware implementation is not possible for the baseline model. The inclusion of batch normalization in their model, in our view, poses a significant challenge on neuromorphic hardware, as it can result in a non-integer number of spikes in the normalized input tensor.

\section{Conclusion} \label{conclusion}
In this paper, we introduced a bio-inspired model to tackle the semantic segmentation task that consists of a spiking light Unet architecture directly fed by an event camera. This architecture has been built using: Surrogate Gradient Learning, Parametric LIF, and SpikingJelly framework. Our semantic segmentation experiments show that our approach outperforms a state-of-the-art model in terms of MIoU and the number of parameters while considering the biological plausibility of the Spiking Neural Network. It also shows a large variation regarding the segmentation result corresponding to each class due to the unbalanced number of pixels belonging to each class. Hence, for future work, it would be interesting to train and test our model using better semantic segmentation labels obtained and generated by e.g. \cite{sun2022ess}. 
Besides, the next step is to implement EvSegSNN on a low-power neuromorphic hardware such as Loihi \cite{orchard2021efficient} or SpiNNaker \cite{furber2020spinnaker} which will enable power-efficient embedded semantic segmentation applications.

\section*{Acknowledgments}

\noindent This work was supported by the project NAMED (ANR-23-CE45-0025-01) of the French National Research Agency (ANR).

\noindent This work was supported by the French government through the France 2030 investment plan managed by the National Research Agency (ANR), as part of the Initiative of Excellence Université Côte d'Azur under reference number ANR- 15-IDEX-01.

\noindent The authors are grateful to the OPAL infrastructure from Université Côte d'Azur for providing resources and support.

{\small
\bibliographystyle{ieee_fullname}
\bibliography{biblio}
}

\end{document}